\documentclass[sigconf]{acmart}

\DeclareUnicodeCharacter{FF0C}{,}

\AtBeginDocument{%
  \providecommand\BibTeX{{%
    \normalfont B\kern-0.5em{\scshape i\kern-0.25em b}\kern-0.8em\TeX}}}

\newcommand\approach{MicroEmo}

\begin{document}

\title{MicroEmo: Time-Sensitive Multimodal Emotion Recognition with Micro-Expression Dynamics in Video Dialogues}



\author{Liyun Zhang}
\affiliation{%
  \institution{Osaka University}
  \country{Japan}
}
\email{liyun.zhang@lab.ime.cmc.osaka-u.ac.jp}



\begin{abstract}
  Multimodal Large Language Models (MLLMs) have demonstrated remarkable multimodal emotion recognition capabilities, integrating multimodal cues from visual, acoustic, and linguistic contexts in the video to recognize human emotional states. However, existing methods ignore capturing local facial features of temporal dynamics of micro-expressions and do not leverage the contextual dependencies of the utterance-aware temporal segments in the video, thereby limiting their expected effectiveness to a certain extent. In this work, we propose \approach, a time-sensitive MLLM aimed at directing attention to the local facial micro-expression dynamics and the contextual dependencies of utterance-aware video clips. Our model incorporates two key architectural contributions: (1) a global-local attention visual encoder that integrates global frame-level timestamp-bound image features with local facial features of temporal dynamics of micro-expressions; (2) an utterance-aware video Q-Former that captures multi-scale and contextual dependencies by generating visual token sequences for each utterance segment and for the entire video then combining them. Preliminary qualitative experiments demonstrate that in a new Explainable Multimodal Emotion Recognition (EMER) task that exploits multi-modal and multi-faceted clues to predict emotions in an open-vocabulary (OV) manner, \approach\ demonstrates its effectiveness compared with the latest methods.  
\end{abstract}












\maketitle

\section{Introduction}
Multimodal Large Language Models (MLLMs) have shown exceptional capabilities in integrating multimodal information (e.g., visual, acoustic, and linguistic modalities) and understanding the context to generate text descriptions \cite{Llama2}. Many MLLMs with outstanding performance can be adapted to different tasks by tuning on specific datasets. However, there still remains a gap in the expected effectiveness. This is because video understanding requires a full perception and deep reasoning of visual/acoustic details and temporal dynamics for models \cite{Video-llama, Vtimellm}, particularly in the multimodal emotion recognition task.

For example, in a new Explainable Multimodal Emotion Recognition (EMER) task \cite{EMER(Multi)} that exploits multi-modal and multi-faceted clues to predict emotions in an open-vocabulary (OV) manner. This not only requires the integration of multimodal information but also demands the in-depth capture of emotion-related subtle information, e.g., micro-expressions. Emotion is a highly sensitive state form that undergoes dynamic changes. The emotions in each utterance of the video evolve dynamically over time, which is crucial information that cannot be overlooked. It necessitates the design of specialized modules within the model to capture and leverage these elements, enabling the provision of detailed recognition of subtle emotional changes without compromising contextual accuracy. However, current existing methods ignore capturing local facial features of temporal dynamics of micro-expressions and do not leverage the contextual dependencies of the utterance-aware temporal segments in the video, thereby limiting their expected effectiveness to a certain extent when tuning MLLMs to Multimodal Emotion Recognition tasks.

To this end, we propose \approach, a novel time-sensitive MLLM aimed at directing attention to the local facial micro-expression dynamics and the contextual dependencies of utterance-aware video clips. \approach\ incorporates two key modules, which consists of \romannumeral1)\, a global-local attention visual encoder that first extracts and integrates global frame-level timestamp-bound image features with local facial features of temporal dynamics of micro-expressions, and \romannumeral2)\, an utterance-aware video Q-Former that that models each utterance time segment feature via an utterance-aware sliding window and combines it with whole features to produce multi-scale fused tokens for the image. 

Benefiting from the attention to local facial features of temporal dynamics of micro-expressions and fully capturing multi-scale and contextual dependencies of utterances in videos, \approach\ shows its superiority in capturing more subtle and deeper emotional information to enable the provision of detailed recognition of subtle emotional changes without compromising contextual accuracy. Finally, audio is also processed by a pre-trained encoder and whole video Q-Former to obtain audio tokens, these multimodal tokens are concatenated with transcribed speech and instructions to feed into the LLM for result generation. Our contributions to this paper are listed as follows:

\begin{itemize}
  \item {We propose a novel global-local attention visual encoder, which integrates global frame-level timestamp-bound image features with local facial features of temporal dynamics of micro-expressions. This facilitates the model to capture more subtle and deeper emotional information to output richer and more reliable results.}
  \item {We propose a novel utterance-aware video Q-Former, which captures multi-scale and contextual dependencies by generating visual token sequences for each utterance segment and for the entire video then combining them. This ensures the accuracy and consistency of contextual dependencies when the model provides detailed recognition of subtle emotional changes.}
  \item {We conduct preliminary qualitative experiments to demonstrate the effectiveness of \approach\ compared with the latest methods in the Explainable Multimodal Emotion Recognition (EMER) task.}
\end{itemize}

\section{Related works}
Multimodal Large Language Model (MLLM) has attracted much attention due to its adaptability to adapt to different tasks by tuning on specific datasets, e.g., tuning on EMER \cite{EMER(Multi)} for multimodal emotion recognition. Numerous studies have endeavored to integrate LLMs with a video encoder for video understanding \cite{Video-Teller, MovieChat, Valley, Timechat}. They typically employ open-source LLMs \cite{Vicuna, LLaMA} to encode the video into vision tokens compatible with the LLMs. Such as VideoChat \cite{Videochat} uses a video transformer to encode video features and then implements a Query Transformer (Q-Former) \cite{BLIP2} for video tokens. Video-llama \cite{Video-llama} uses a vision transformer (ViT) \cite{ViT} with an image Q-Former to encode frames and employs a video Q-Former for temporal modeling. TimeChat \cite{Timechat} incorporates frame timestamps to make the model time-sensitive and uses a video Q-Former to establish inter-frame temporal dependencies.

However, these methods ignore capturing local features of temporal dynamics and do not leverage the contextual dependencies of important segments in videos, thereby limiting their expected effectiveness to a certain extent. In contrast, our \approach\ proposes a global-local attention visual encoder and an utterance-aware video Q-Former to capture subtle local features of temporal dynamics and contextual dependencies in videos, such as micro-expressions and utterance segment context of emotion recognition, enabling the provision of subtle reasoning results without compromising contextual accuracy.

\begin{figure*}[ht]
\begin{center}
\includegraphics[width=1.0\linewidth]{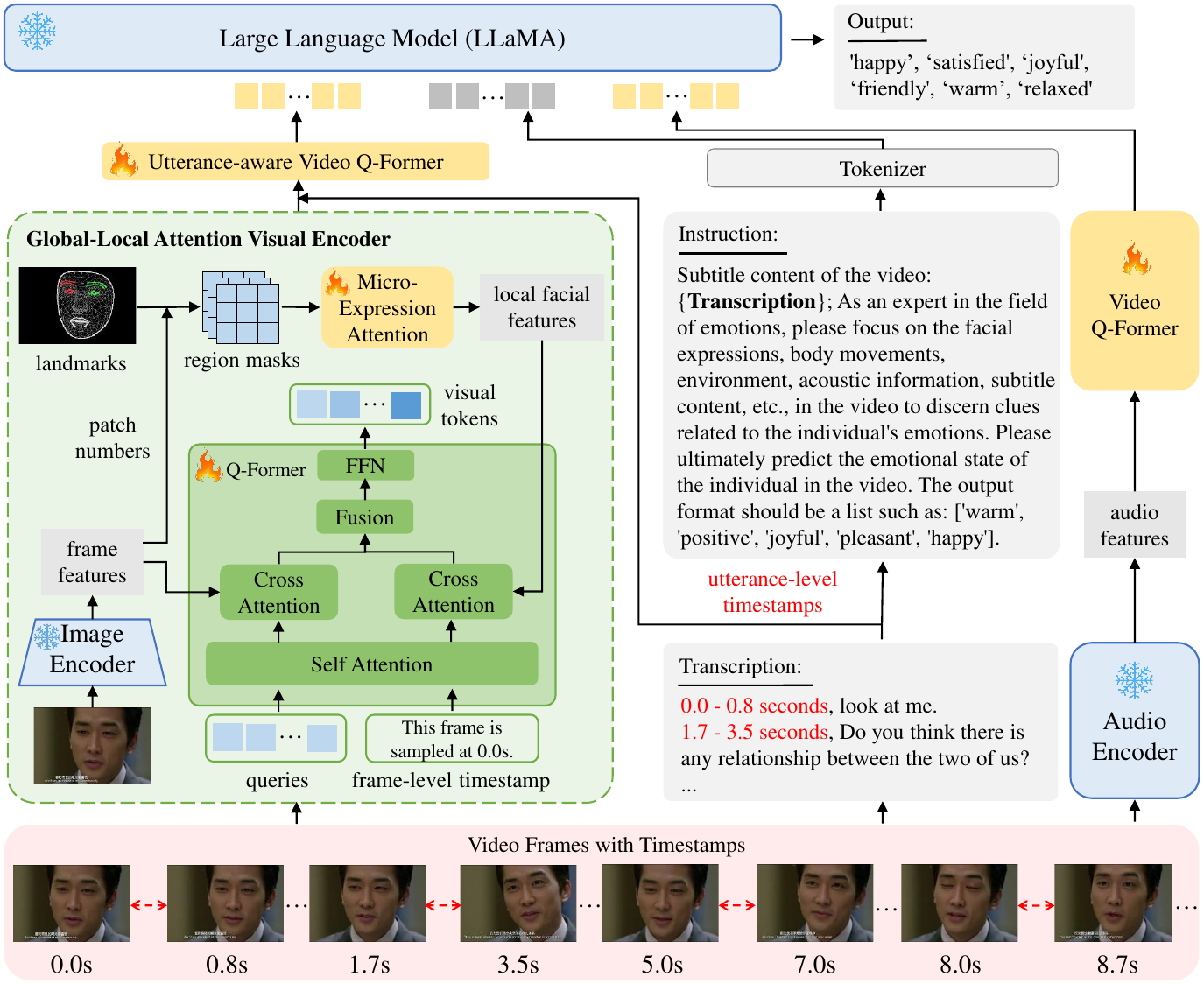}
\end{center}
   \caption{The overall architecture of \approach. Input a sequence of video frames along with their timestamps, (a) Global-local attention visual encoder first extracts and integrates global frame-level timestamp-bound image features with local facial features of temporal dynamics of micro-expressions. Then (b) utterance-aware video Q-Former combines utterance-level timestamps to generate visual tokens for each utterance segment via an utterance-aware sliding window and combines them with whole tokens extracted by a global video Q-Former to produce multi-scale fused visual tokens. Audio is also passed through a pre-trained encoder and whole video Q-Former to obtain audio tokens. Finally, these multimodal tokens are concatenated with transcribed speech and instructions, which are then fed into a (c) Large Language Model to generate responses.}
\label{fig:framework}
\end{figure*}

\section{Methodology}
We present \approach, a time-sensitive MLLM featuring two novel modules: a global-local attention visual encoder and an utterance-aware video Q-Former. These modules enable \approach\ to capture more subtle and deeper emotional information in videos. We verify the effectiveness of our model on the new task EMER and the new style dataset EMER-Fine \cite{EMER(Multi)} it pioneered.

\subsection{Overview}
\approach\ is mainly composed of a global-local attention visual encoder, an utterance-aware video Q-Former for image, and a large language model, as depicted in Figure \ref{fig:framework}. Given an input video, the global-local attention visual encoder first extracts and integrates global frame-level timestamp-bound image features with local facial features of temporal dynamics of micro-expressions. Next, the utterance-aware video Q-Former models each utterance time segment feature via an utterance-aware sliding window and combines it with whole features to produce multi-scale fused tokens for the image. Audio is also passed through a pre-trained encoder and whole video Q-Former to obtain audio tokens. Finally, these multimodal tokens are concatenated with transcribed speech and instructions, which are then fed into the LLM to generate responses.

\subsection{Global-Local Attention Visual Encoder}
First, for the global frame-level features, we use the time-aware frame encoder inspired by InstructBLIP \cite{InstructBLIP} and Timechat \cite{Timechat} as shown in Figure \ref{fig:framework}. Given an input video, we use a pre-trained ViT \cite{EVA-CLIP} to encode each frame and obtain frame features. Subsequently, we use the keypoint indices of the MediaPipe \cite{MediaPipe} face mesh via extracting landmarks of each frame to define different facial regions as micro-expression regions, which are predefined on facial anatomy. 
Then, the patches are combined to create masks associated with micro-expression regions. we implemented a localized micro-expression attention mechanism based on the generated mask. This allows each patch to selectively attend to pertinent micro-expression regions, facilitating the extraction of fine-grained facial local features. By emphasizing spatially relevant areas, we enhance the model's capacity to discern subtle facial cues and micro-expressions, which are crucial for accurate emotion recognition.

Furthermore, the image Q-Former compresses the frame tokens. The Q-Former takes learnable queries as input, which interacts with the frame features and facial local features via cross-attention and updates the initial queries respectively. The global and local results are fused and further processed to obtain final visual tokens \cite{BLIP2}.
During visual token extraction, we add the frame-level timestamp, e.g., \texttt{"This frame is sampled at 0.0s."}, as a condition to the Q-Former to fuse the visual and timestamp information for binding temporal dynamics to the visual tokens.

The proposed global-local attention visual encoder can fully integrate global timestamp-bound frame features with local facial features of temporal dynamics of micro-expressions.
Benefiting from the attention to temporal dynamics of micro-expressions of the videos, \approach\ shows its superiority in capturing more subtle and deeper emotional information to facilitate richer and more reliable emotion recognition results.

\subsection{Utterance-Aware Video Q-Former}
Since frames are encoded independently, we can use a sliding video Q-Former to model temporal relationships across frames and enhance the feature fusion in the temporal dimension \cite{Timechat}. Here, we propose a novel utterance-aware video Q-Former (yellow block in Figure \ref{fig:framework}), which combines utterance-level timestamps to design a dynamic sliding window with a length of each utterance segment and within each window utilizing the video Q-Former extract video tokens from frame features. Meanwhile, we also extract video tokens from whole frame features via a global video Q-Former and then combine them to produce multi-scale fused tokens.

It is crucial to capture the context of each conversation segment, which is the key to the model's understanding of emotional fluctuations and changes. We use this way to fully capture multi-scale and contextual dependencies, maintaining the accuracy and consistency of contextual dependencies when the model provides detailed recognition of subtle emotional changes. Notably, the method maintains its effectiveness even in videos with minimal or no dialogue content, thus ensuring robustness across diverse types of video material. Finally, we use a linear layer to transform the dimension of video tokens to match the dimension of the LLM embedding space.

\subsection{Large Language Model}
We concatenate inputs from different modalities, i.e., the video tokens $T_V$, audio tokens $T_A$,  and text query tokens $T_{L_q}$ including transcribed speech and instruction. They are fed into a large language model to generate reasonable and expected responses $T_{L_a}$. They have the same token embedding dimension. The training of the MLLM typically utilizes a two-stage training framework. The first stage pre-trains the model using large-scale image-audio-text pairs for vision-audio-language mapping \cite{VITATECS, FETV, Delving, VideoCLIP}. The second stage finetunes the model with instruction data for instruction following. Considering computing efficiency, we reuse the checkpoints of the existing open-source models after the first stage of training, e.g., the vision-language and audio-language branches from \cite{Video-llama}, conducting only instruction tuning. During the training procedure, we utilize the language modeling loss for generating target answers $T_{L_a}$ and $\Theta$ is the trainable parameter, which serves as the objective function:

\begin{equation}
  L = - log  P_{\Theta}(T_{L_a} | T_{V}, T_{A}, T_{L_q})
\end{equation}

To better adapt the LLM to multimodal video tasks, we apply the parameter-efficient fine-tuning method, i.e., LoRA \cite{LoRA} to train.

\section{Experiments}

\subsection{Implementation Details}
We use ViT-G/14 from EVA-CLIP \cite{EVA-CLIP} as the image encoder, Imagebind \cite{ImageBind} as the audio encoder, and LLaMA-2 (7B) \cite{Llama2} as the language foundation model. The parameters of the image Q-Former are initialized from InstructBLIP's checkpoint, while the video Q-Former for vision and audio is initialized from Video-LLaMA's checkpoint.  We finetune our \approach\ on train-set of EMER-Fine \cite{EMER(Multi)} dataset for 3 epochs, using a batch size of 1, with a single NVIDIA A40 (49G) machine. The parameters of ViT, Imagebind and LLM are frozen, while those of image Q-Former, micro-expression attention module, video Q-Former, and linear layer (not shown in the diagram due to space constraints) are tuned.

\subsection{Experiment Setup}
{\bf Tasks, Datasets and Evaluation Metrics} \hspace{0.15cm} We evaluate our model on a new Explainable Multimodal Emotion Recognition (EMER) task \cite{EMER(Multi)}, test-set of EMER-Fine as the test dataset, the new $Accuracy_{s}$ and $Recall_{s}$ in EMER as metrics, and MLLM such as VideoChat \cite{Videochat}, Video-LLaMA \cite{Video-llama} and Chat-UniVi \cite{Chat-univi}, etc. as baselines for evaluation.

\begin{table}[h]
\caption{Main results on the test-set of EMER-Fine \cite{EMER(Multi)}.}
\label{tab:main_results}
\begin{tabular}{lccc}
\toprule
Model & Avg & Accuracy$_S$ & Recall$_S$ \\
\midrule
\multicolumn{4}{c}{Audio + Subtitle} \\
\midrule
Qwen-Audio \cite{Qwen-audio} & 38.66 & 46.97 & 30.35 \\
OneLLM \cite{Onellm} & 40.56 & 42.55 & 38.56 \\
SECap \cite{Secap} & 52.78 & 61.36 & 44.19 \\
SALMONN \cite{Salmonn} & 51.28 & 54.17 & 48.38 \\
\midrule
\multicolumn{4}{c}{Video + Subtitle} \\
\midrule
Otter \cite{Otter} & 37.72 & 42.22 & 33.22 \\
VideoChat \cite{Videochat} & 46.34 & 41.49 & 51.19 \\
Video-LLaMA \cite{Video-llama} & 40.97 & 39.44 & 42.50 \\
Video-LLaVA \cite{Video-llava} & 42.75 & 45.38 & 40.13 \\
VideoChat2 \cite{VideoChat2} & 43.83 & 49.24 & 38.42 \\
OneLLM \cite{Onellm} & 53.40 & 58.65 & 48.14 \\
LLaMA-VID \cite{Llama-vid} & 50.90 & 51.69 & 50.11 \\
mPLUG-Owl \cite{mplug-owl} & 48.84 & 48.33 & 49.34 \\
Video-ChatGPT \cite{Video-chatgpt} & 46.12 & 50.00 & 42.25 \\
Chat-UniVi \cite{Chat-univi} & 53.20 & 54.29 & 52.11 \\
\midrule
\multicolumn{4}{c}{Audio + Video + Subtitle} \\
\midrule
SECap + mPLUG-Owl & 64.42 & 57.95 & 70.90 \\
SALMONN + Video-ChatGPT & 59.71 & 53.48 & 65.93 \\
SECap + Video-ChatGPT & 58.43 & 51.60 & 65.26 \\
SECap + Chat-UniVi & 60.38 & 51.39 & 69.37 \\
SALMONN + mPLUG-Owl & 65.15 & 55.28 & \bf{75.03} \\
SALMONN + Chat-UniVi & 62.64 & 55.84 & 69.44 \\
AffectGPT \cite{EMER(Multi)} & 61.75 & 62.03 & 61.46 \\
Ours & \bf{66.21} & \bf{63.82} & 68.59 \\
\midrule
EMER (Multi) \cite{EMER(Multi)} & 79.31 & 80.91 & 77.70 \\
\bottomrule
\end{tabular}
\end{table}

\subsection{Main Results}
We evaluate our \approach\ and some existing MLLMs for EMER task as shown in Table~\ref{tab:main_results}, where, the baseline methods except for AffectGPT \cite{EMER(Multi)} are zero-shot performance results. AffectGPT is only tuned on the EMER-Fine \cite{EMER(Multi)} dataset in the second stage without pre-training of the first stage,  i.e., pre-training the model using large-scale image-audio-text pairs for vision-audio-language mapping, which is the same as our experimental setting. EMER (Multi) \cite{EMER(Multi)} can be viewed as a post-processed upper bound approximating the ground truth, serving as a reference point here. The experimental results indicate that our proposed model outperforms the baseline methods overall, demonstrating the superiority and effectiveness of our approach. Incorporating large-scale multimodal data alignment pre-training in the first stage is expected to result in further performance improvements.

\begin{table}
\caption{Ablation study for key modules.}
\label{tab:ablation}
\begin{tabular}{lccc}
\toprule
Model & Avg & Accuracy$_S$ & Recall$_S$ \\
\midrule
\approach\ & 66.21 & 63.82 & 68.59 \\
w/o utterance-aware V-QF
 & 56.01 & 52.53 & 59.49 \\
w/o global-local AVE 
& 54.57 & 50.43 & 58.71 \\
w/o all proposed modules & 46.21 & 49.75 & 47.98 \\
\bottomrule
\end{tabular}
\end{table}

\subsection{Ablation Study}
We conducted an ablation study to evaluate the key novel modules, i.e., the global-local attention visual encoder (AVE) and utterance-
aware video Q-Former (V-QF). As shown in Table~\ref{tab:ablation}, removing either of these modules decreases the overall performance, highlighting their effectiveness and reasonability in the architecture.

\section{Conclusion}
We propose \approach\ that directs attention to the local facial micro-expression dynamics and fully captures multi-scale and contextual dependencies of utterances in videos, and tuning on EMER. \approach\ demonstrates capabilities in capturing subtle emotion to output richer and more reliable results without compromising contextual accuracy. In the future, we will apply the utterance-aware video Q-Former to acoustic modality and integrate emotion in audio into multimodal emotion recognition task based on MLLM.




\bibliographystyle{ACM-Reference-Format}
\bibliography{sample-base}

\end{document}